\pdfoutput=1
\documentclass[11pt]{article}

\usepackage[final]{acl}

\usepackage{times}
\usepackage{latexsym}

\usepackage[T1]{fontenc}
\usepackage[utf8]{inputenc}

\usepackage{microtype}

\usepackage{inconsolata}

\usepackage{graphicx}
\usepackage{array}
\usepackage{multirow}
\usepackage{booktabs}
\usepackage{tabularx}
\usepackage{amsmath}

\usepackage{subcaption}
\usepackage{enumitem}
\title{Evaluating the Performance of RAG Methods for Conversational AI in the Airport Domain}

\author{
 \textbf{Yuyang Li\textsuperscript{1}},
 \textbf{Philip J.M. Kerbusch\textsuperscript{2}},
 \textbf{Raimon H.R. Pruim\textsuperscript{2}},
 \textbf{Tobias Käfer\textsuperscript{1}}
\\
\\
 \textsuperscript{1}Karlsruhe Institute of Technology,
 \textsuperscript{2}Royal Schiphol Group \\
 \textsuperscript{2}Royal Schiphol Group,
 \textsuperscript{1}Karlsruhe Institute of Technology
\\
  \texttt{yuyang.li@kit.edu}, \\
  \texttt{tobias.kaefer@kit.edu}}

\begin{document}
\maketitle
\begin{abstract}
% The airport is a highly dynamic environment with over 1,500 flights daily, and it aims to become autonomous. To achieve this, we implemented a Conversational AI system that enables airport staff to communicate with flight information systems. This conversational AI not only answers standard airport queries but also resolves ambiguous, airport-specific questions. We generated two benchmark datasets from flight data to evaluate different methods: a straightforward questions dataset and a complicated/ambiguous questions dataset, each containing thousands of questions and answer pairs. Different Retrieval-Augmented Generation (RAG) pipelines were developed, including traditional RAG (using the keyword or semantic search with Large Language Models), SQL RAG, and Knowledge Graph-based RAG (Graph RAG). Experiments showed that traditional RAG achieved 84.84\% accuracy using BM25 + GPT-4 on the straightforward dataset but occasionally produced hallucinations, especially on complex and ambiguous queries (up to 10\%), which is risky to airline security. In contrast, SQL RAG and Graph RAG proved more suitable for the airport environment. SQL RAG and Knowledge Graph RAG achieved 80.85\% and 91.49\% accuracy respectively, with significantly fewer hallucinations. Furthermore, Graph RAG achieved 68.75\% accuracy on questions requiring reasoning, compared to 6.25\% and 9.38\% for SQL RAG and traditional RAG, respectively.

Airports from the top 20 in terms of annual passengers are highly dynamic environments with thousands of flights daily, and they aim to increase the degree of automation. To contribute to this, we implemented a Conversational AI system that enables staff in an airport to communicate with flight information systems. This system not only answers standard airport queries but also resolves airport terminology, jargon, abbreviations, and dynamic questions involving reasoning. In this paper, we built three different Retrieval-Augmented Generation (RAG) methods, including traditional RAG, SQL RAG, and Knowledge Graph-based RAG (Graph RAG). Experiments showed that traditional RAG achieved 84.84\% accuracy using BM25 + GPT-4 but occasionally produced hallucinations, which is risky to airport safety. In contrast, SQL RAG and Graph RAG achieved 80.85\% and 91.49\% accuracy respectively, with significantly fewer hallucinations. Moreover, Graph RAG was especially effective for questions that involved reasoning. Based on our observations, we thus recommend SQL RAG and Graph RAG are better for airport environments, due to fewer hallucinations and the ability to handle dynamic questions.

% top 20 airports have 56 million passengers per year, 104 million passengers

\end{abstract}

\section{Introduction}
Amsterdam Airport Schiphol, one of the top 20 airports in the world, ranked by annual passenger numbers, handles thousands of flights each day. These airports rely on staff like gate planners and apron controllers to access and update data across systems. For these employees, traditional database queries can be complex and time-consuming for some employees who are not query experts when they need flight information. A conversational AI system with a natural language query (NLQ) interface allows all employees to interact with systems naturally, asking questions like, “Which fights are at ramp D07?” and receiving instant answers. This improves productivity, and streamlines workflows, especially in high-pressure areas like at the gate, where less educated workers require access to up-to-date information.
By replacing strict query formats with intuitive, real-time responses, conversational AI enhances decision-making and efficiency, making it a suitable solution for dynamic environments such as airports.

Building such a system is challenging because flight data is stored by experts in tables using aviation abbreviations. We need our system to understand these datasets to answer questions from the airport domain. Additionally, ensuring aviation safety is a major concern; the system must be safe and enable employees to perform accurate operations.
We address those challenges using two research questions.

The first question is how to handle flight data so that our system can answer different questions. We divided the questions into three types: 
\begin{itemize}
        \item \textbf{Straightforward questions:} Questions that can be directly answered from the flight data.
        
        \item \textbf{Questions involving specialized airport jargon, abbreviations, and incomplete queries:} 
        Operators often use shorthand or omit context. Flight ``KL0123'' might be referred to as ``0123'' or ``123,'' while gate ``C05'' might be shortened to ``C5.'' Abbreviations like ``KLM'' for ``KLM Royal Dutch Airlines'' or ``Delta'' for ``Delta Air Lines'' are also common. Operators frequently ask short, incomplete questions, e.\,g., ``Which flights are at D04?'' or ``What is the gate for that Delta airline?'' Without resolving missing details such, these questions cannot be answered.

        \item \textbf{Dynamic questions:} Questions that involve additional calculations and reasoning, especially related to time. Examples include ``What is the connecting flight's onramp time for DL1000?'' or ``What is DL1000's next flight from the same ramp?'' These queries require reasoning through connections between flights and retrieving specific details.
    \end{itemize}

The second research question is about how to reduce hallucinations \cite{xu2024hallucinationinevitableinnatelimitation} for the safety of aviation operations. Hallucinations occur when LLMs generate information not based on facts or their training data. In high-safety environments such as airports, however the output should be factual and not imaginative~\cite{jacobs2024leveraging}. For example, if the system gives wrong gate numbers, flight schedules, or safety instructions, this might disrupt aviation operations, cause delays, or even risk passenger safety. Thus, accurate responses are important.

In this case study, we examine three Retrieval-Augmented Generation (RAG) techniques for the airport environment:
Traditional RAG  \cite{lewis2021retrievalaugmentedgenerationknowledgeintensivenlp} Retrieves relevant information from the flight database and uses LLMs to generate answers based on the retrieved data and original questions. SQL RAG \cite{guo2023retrievalaugmentedgpt35basedtexttosqlframework} stores all datasets in an SQL database and converts natural language questions (NLQ) into structured SQL queries. Knowledge Graph-based Retrieval-Augmented Generation (Graph RAG) \cite{edge2024localglobalgraphrag} aims to improve the performance of LLM tasks by applying RAG techniques to Knowledge Graphs (KGs), requiring the original datasets to be stored in the knowledge graph. A key challenge is retrieving the correct flight information from thousands of flights while minimizing hallucinations. 

The paper is structured as follows: We first survey related work (Sec.~\ref{sec:related-work}), then present our dataset (Sec.~\ref{sec:dataset}), followed by a high-level description of our experiments (Sec.~\ref{sec:experiments}).
We then present the results for the research questions (Sec.~\ref{sec:results}), and lastly conclude (Sec.~\ref{sec:conclusion}).
In the Appendix~\ref{sec:appendix}, we provide further details, especially on the question generation and classification, next to our prompting.

\section{Related Work}
\label{sec:related-work}
\subsection{Traditional RAG}

Traditional Retrieval-Augmented Generation (RAG) consists of two main stages: the Retriever and the Generator 
\cite{louis2023interpretable}. The Retriever identifies relevant documents based on user input, and the Generator uses these documents to produce responses. We explore three retrieval methods: keyword search, semantic search, and hybrid search, using large language models (LLMs) for answer generation.

In keyword search, TF-IDF and BM25 are employed to evaluate retrieval performance. TF-IDF computes term frequency (TF) and inverse document frequency (IDF) \cite{8492945, article}, measuring how important a term is within a document and across the corpus. BM25 extends TF-IDF with a term saturation function \cite{article2}, reducing the influence of extremely frequent terms that often carry less informative value \cite{chen2023bm25queryaugmentationlearned}.

Semantic search methods include similarity search, vector databases like FAISS \cite{faiss2017, 10039758}, k-Nearest Neighbors (KNN), Locality-Sensitive Hashing (LSH) \cite{jafari2021surveylocalitysensitivehashing}, and Maximal Marginal Relevance (MMR) \cite{mao2020multidocumentsummarizationmaximalmarginal}. Unlike keyword search, semantic search aims to understand user intent and word meanings \cite{gao2024retrievalaugmented}. Embedding models such as Word2Vec convert words into vectors \cite{mikolov2013efficientestimationwordrepresentations}, where cosine similarity measures similarities between queries and documents.

Hybrid search combines keyword and semantic methods, re-ranking results using the Reciprocal Rank Fusion (RRF) algorithm \cite{microsoft_azure}. By combining two search methods, the hybrid search can not only find flight information by keywords but also find information by the deeper meaning of the queries \cite{sarmah2024hybridragintegratingknowledgegraphs}.

% The Ensemble Retriever in the Langchain framework is used to merge results from multiple retrievers, 

\subsection{SQL RAG}
Text-to-SQL aims to transfer natural language automatically
questions(NLQs) into SQL queries. LLMs recently emerged as an option for Text-to-SQL task \cite{rajkumar2022evaluatingtexttosqlcapabilitieslarge}. The trick to handling text-to-SQL tasks with LLMs is to apply prompt engineering. Five prompt styles for Text-to-SQL are explored in the previous research \cite{gao2023texttosqlempoweredlargelanguage}. Basic Prompt (BSP) is a simple representation with no instructions; Text Representation Prompt (TRP) adds basic task guidance; OpenAI Demonstration Prompt (ODP) adds explicit rules like “Complete sqlite SQL query only,”; Code Representation Prompt (CRP) uses SQL-style schema descriptions with detailed database information like primary/foreign keys, and Alpaca SFT Prompt (ASP) adopts Markdown for structured training prompts. In \cite{gao2023texttosqlempoweredlargelanguage}, CRP achieves the best performance in most LLMs, by providing complete database information and utilizing the LLMs’ strength in understanding code.

\subsection{Graph RAG}

A Knowledge Graph (KG) is a structured representation of entities (nodes), their attributes, and relationships (edges), typically stored in graph databases or triple stores \cite{sarmah2024hybridragintegratingknowledgegraphs}. Its basic unit is a triple: subject, predicate, object. In Graph RAG (Retrieval-Augmented Generation), natural language questions are converted into query languages like SPARQL for RDF graphs or Cypher for Neo4j property graphs. Research indicates that Neo4j's labeled property graph model offers faster and more efficient real-time analysis and dynamic querying compared to complex RDF ontologies in enterprise projects \cite{barrasa2023building}. Neo4j's property graph model better meets industrial needs. Flight information can be automatically integrated into the knowledge graph by matching the row and column names of the flight table, with relationships manually defined based on flight numbers.
    
\section{Dataset and Questions}
\label{sec:dataset}
 Our flight information dataset is tabular containing thousands of flights with key details such as flight number, aircraft category, bus gate, bus service needed, flight UID, ramp, expected on-ramp time, connecting flight number, etc.

To evaluate the effectiveness of different retrieval methods, we classified the questions, and then based on these questions, we created two ground truth datasets: a straightforward dataset and a complicated, ambiguous dataset.

The straightforward dataset consists of unambiguous questions that can be directly answered from flight information. Examples include: "What category of aircraft is designated for flight KL1000?" and "Which ramp is assigned for flight KL1000?". Such questions are easily handled by retrieval methods to select the most relevant information. This dataset contains thousands of question-answer pairs, with around 100 to 200 pairs selected for the RAG methods comparison. 

The complicated and ambiguous dataset contains questions with variables that may be unclear or missing from the flight information which cannot be directly queried from the tabular dataset. Examples are: "Which flight is at gate B24?" or "Which gate is assigned to the 0164 flight?", "When is Delta landing?" Here, 'B24' might relate to multiple flights or meanings (bus gate or ramp number), and '0164' is not a complete flight number, 'Delta' also needs clarification. This dataset also contains thousands of question-answer pairs, with 185 pairs randomly selected for comparison. More information on question generation and question classification is provided in the Appendix \ref{sec:appendix}.

\section{Experiments}
\label{sec:experiments}

To handle the flight tabular dataset, our conversational AI should understand the meaning of these flight terms, it also needs to understand specific jargon and terminology. We explore three RAG methods for a conversational system on flight data.

\begin{figure}[h!]
  \vspace{-5pt} % 减少图片上方空白
  \includegraphics[width=0.9\columnwidth]{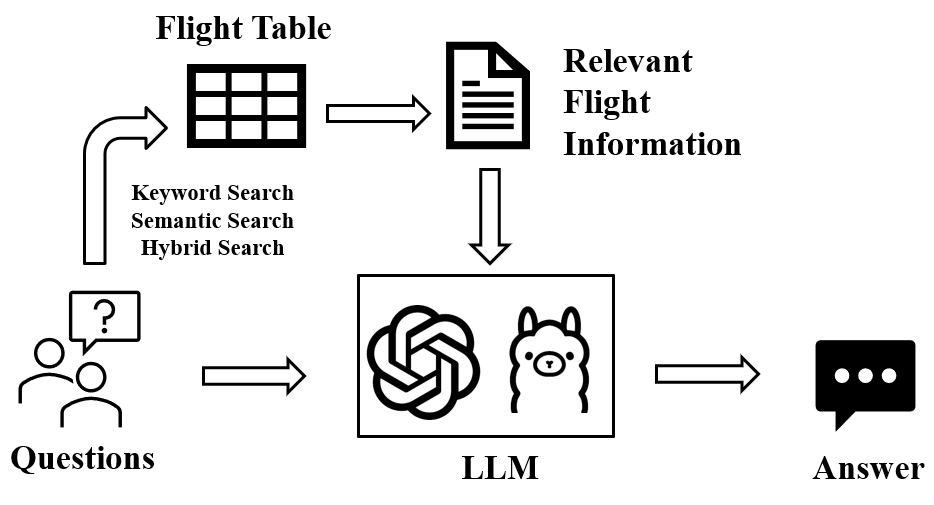}
  \vspace{-10pt} % 减少图片下方空白
  \caption{Traditional RAG Method}
  \label{fig:tradtional_rag}
\end{figure}

Figure \ref{fig:tradtional_rag} shows the traditional RAG method. When a user asks a question, various retrieval methods are employed to retrieve the correct flight data from the flight information dataset. These methods are mainly divided into three categories: keyword search, semantic search, and hybrid search. After retrieving the relevant flight information, Large Language Models (LLMs) generate answers to the user's questions based on this data. Several LLMs were tested to assess their performance, including GPT models, Llama-3-8B-Instruct, BERT, and BERT-related models.

\begin{figure}[h!]
  \vspace{-5pt}
  \includegraphics[width=0.9\columnwidth]{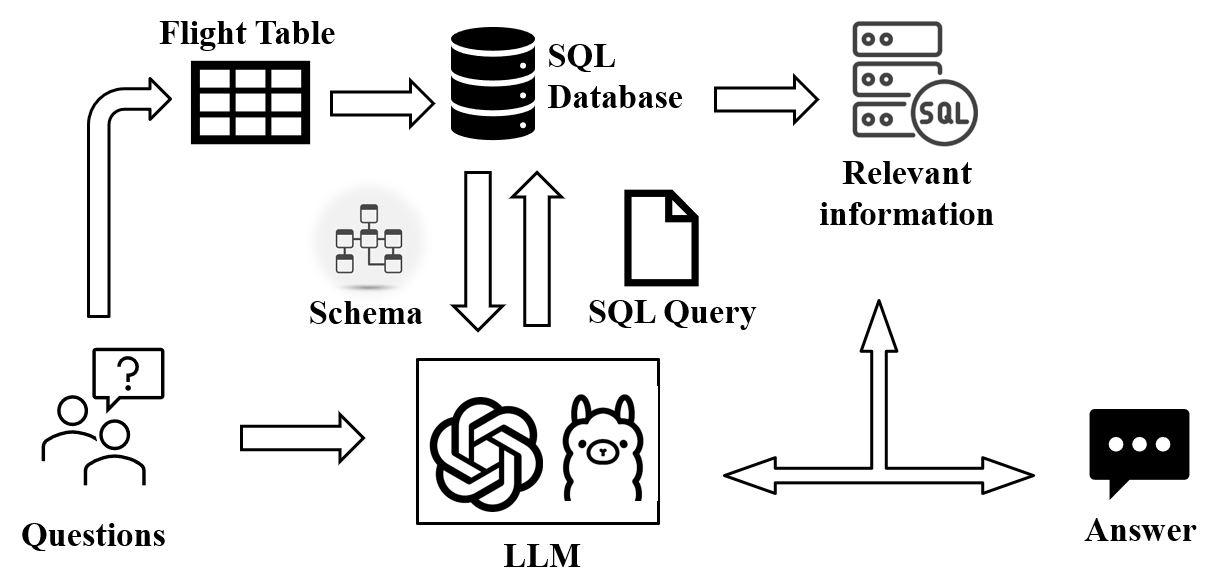}
  \vspace{-10pt}
  \caption{SQL RAG Method}
  \label{fig:SQL_rag}
\end{figure}

Figure \ref{fig:SQL_rag} shows the SQL RAG method, which begins with users asking natural language questions. An LLM processes these questions using the SQL database schema to generate appropriate SQL queries. The queries retrieve relevant information from the SQL database, which the LLM then interprets and reformulates into human-readable answers. Following the approach in \cite{gao2023texttosqlempoweredlargelanguage}, we experimented with Code Representation Prompt (CRP) and OpenAI Demonstration Prompt (ODP) to fine-tune the prompts and improve the SQL RAG results.  More details of SQL RAG prompts are provided in Appendix~\ref{sec:appendix}.

Figure \ref{fig:Graph_rag} shows the Graph RAG method, which also starts with users asking natural language questions. An LLM processes these questions using the graph schema from the graph database to generate graph queries. We use Neo4j's APOC plugin to extract the schema by executing 'CALL apoc.meta.schema() YIELD value RETURN value' and include it in the prompt. and the LLM interprets this data to formulate human-readable answers. The graph structure enables context-aware retrieval and reasoning, more details of Graph RAG prompts are provided in Appendix~\ref{sec:appendix}.

\begin{figure}[h!]
  \vspace{-5pt}
  \includegraphics[width=0.9\columnwidth]{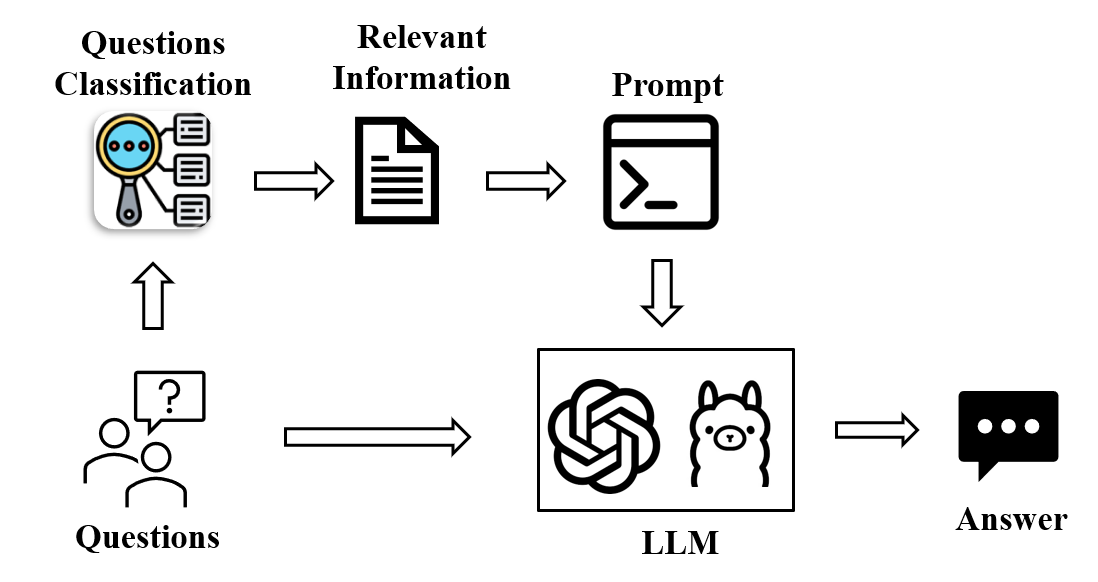}
  \vspace{-10pt}
  \caption{Method on Ambiguous question dataset}
  \label{fig:QA_class_pipeline}
\end{figure}

The three RAG methods described above can handle straightforward datasets easily because the answers all exist in the flight tabular, we will add some explanations about flight row names' meanings to the prompts, so that LLMs can generate better more accurate answers. However, questions about jargon and short sentences from complicated or ambiguous datasets need to be classified using a question classification prompt, as shown in Figure \ref{fig:QA_class_pipeline}. After classification, each question is directed to different prompts to answer jargon and abbreviations.

\begin{figure}[h!]
  \vspace{-5pt}
  \includegraphics[width=0.9\columnwidth]{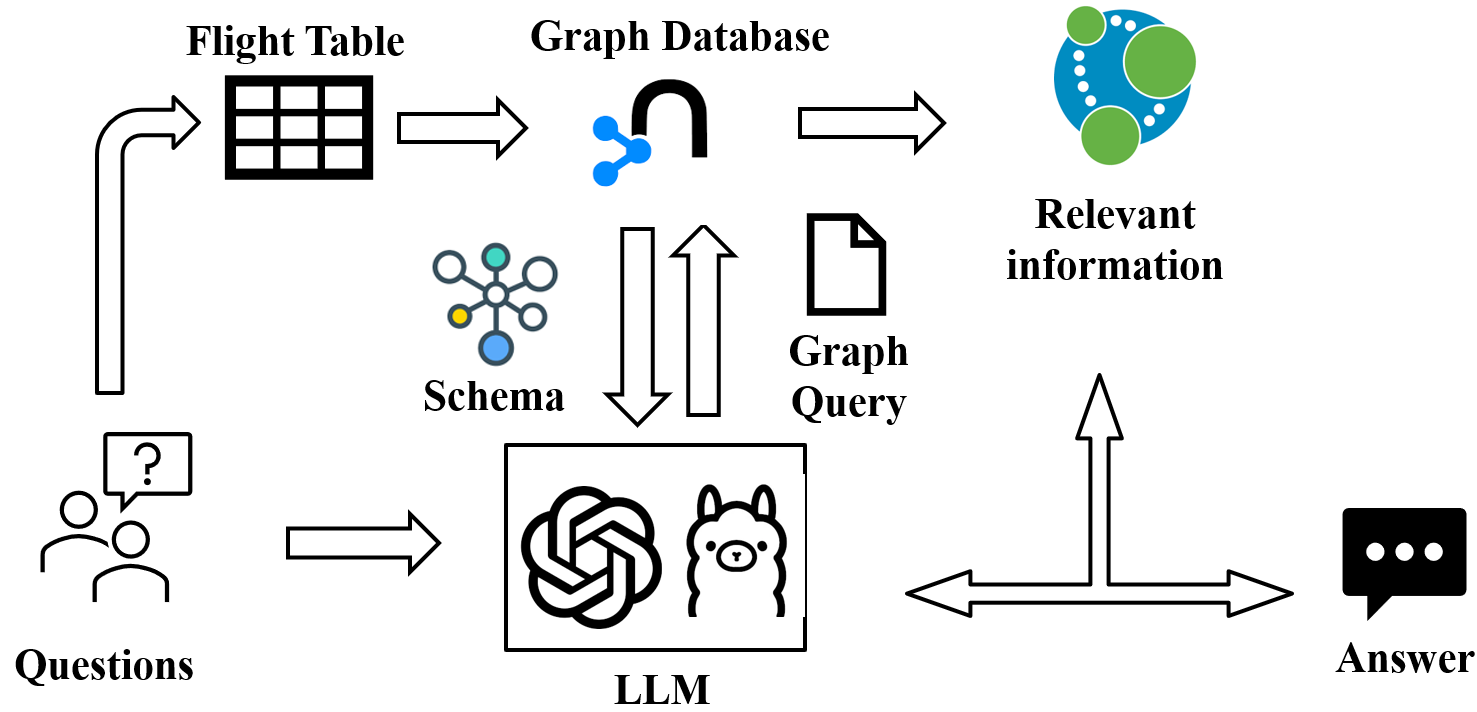}
  \vspace{-10pt}
  \caption{Graph RAG Method}
  \label{fig:Graph_rag}
\end{figure}

\section{Results}
\label{sec:results}
In this section, we present the experimental results, structured using our research questions.

\subsection{RQ1: How to handle flight data for different questions?}

\subsubsection{Straightforward questions}

Table \ref{tab:static_dataset_results} summarizes the performance of various retrieval methods within the traditional RAG pipeline in the straightforward dataset. BM25 outperforms other methods, achieving approximately 86.54\% accuracy in retrieving the correct articles. The hybrid search, which combines BM25 and the vector database FAISS in a 9:1 proportion, performs second best, with an accuracy of 85.78\% for identifying the correct article as the highest-ranked and 98.00\% accuracy for including the correct article within the top 10 results. This indicates the successful retrieval of correct articles among the top 10 most relevant ones. However, changing the proportion to 1:9 yields only 0.59\% accuracy within the top 30 articles, suggesting that the correct articles rarely appear among the top 30 results. Following BM25 and the hybrid search, TF-IDF with cosine similarity and Euclidean distance achieve accuracies of 67.70\% and 67.55\%, respectively. The vector database FAISS alone performs the worst, with an accuracy of 0\%.

Table \ref{tab:generators} shows how various LLMs perform in generating answers for the simple dataset. Because the dataset is large, we randomly selected 100 questions for the experiment. The LLMs' answers were manually compared to the standard answers; correct ones were marked "True," and incorrect ones were "False." Accuracy was calculated by dividing the number of correct answers by the total number of questions. In these two tables, we chose BM25+GPT4 as the traditional RAG pipeline and achieved a total accuracy of 84.40\% in the end. The reason keyword search outperforms semantic search is probably because, in the airport environment, most questions are about specific flights, times, or ramps. These questions don't require a deep semantic understanding of the content.

% Adjusting the proportion of BM25 and FAISS to 5:5 results in an accuracy of 82.37\% for the highest-ranked articles.  Semantic similarity searches using Word2Vec with cosine similarity and MMR achieve an accuracy of 33.70\%, while semantic search with LSI achieves 21.82\%. 

\renewcommand{\arraystretch}{0.9} % 缩小行距
\setlength{\tabcolsep}{3pt} % 缩小列间距

\newcolumntype{C}[1]{>{\centering\arraybackslash}p{#1}}

\begin{table*}[h!]
\centering
\small
\begin{tabular}{C{6.5cm} C{1.8cm} C{2cm} C{2cm} C{2cm}}
\hline
\textbf{Retrieval Methods} & \textbf{Total Rows} & \textbf{Accuracy (Highest)} & \textbf{Accuracy (Top 10)} & \textbf{Accuracy (Top 30)} \\
\hline
BM25 & 1350 & 86.54\% & 100\% & 100\% \\
TF-IDF + Cosine Similarity & 1350 & 67.70\% & 100\% & 100\% \\
TF-IDF + Euclidean Distance & 1350 & 67.55\% & 100\% & 100\% \\
Word2Vec + Cosine Similarity + MMR & 1350 & 33.70\% & 34.00\% & 34.00\% \\
LSI & 1350 & 21.82\% & 37.00\% & 45.00\% \\
FAISS & 1350 & 0\% & 1.00\% & 12.00\% \\
Hybrid Search (BM25 : FAISS = 9:1) & 1350 & 85.78\% & 98.00\% & 98.00\% \\
Hybrid Search (BM25 : FAISS = 5:5) & 1350 & 82.37\% & 98.00\% & 98.00\% \\
Hybrid Search (BM25 : FAISS = 1:9) & 1350 & 0.59\% & 0.59\% & 0.59\% \\
\hline
\end{tabular}
\caption{Retrieval method results for the Traditional RAG in the straightforward dataset}
\label{tab:static_dataset_results}
\end{table*}

\begin{table}[h!]
\centering
\small
\begin{tabular}{l c}
\hline
\textbf{Model Name} & \textbf{Accuracy} \\
\hline
GPT-4           & 88.78\% \\
GPT-4o Mini      & 88.12\% \\
GPT-3.5 Turbo   & 83.33\% \\
Llama-3-8B-Instruct         & 76.54\% \\
RoBERTa         & 56.16\% \\
BERT            & 29.73\% \\
DistilBERT      & 28.00\% \\
DeBERTa         & 41.89\% \\
mDeBERTa        & 53.33\% \\
Electra         & 41.33\% \\
Electra Large   & 41.33\% \\
\hline
\end{tabular}
\caption{LLMs results in straightforward dataset}
\label{tab:generators}
\end{table}

Table \ref{tab:sql_rag} shows the performance of SQL RAG. The results indicate that CRP significantly outperforms ODP in most of the cases. EM(Exact Match) measures the strict match between the predicted SQL query and the ground truth regarding syntax and structure. while EX(Execution Match) evaluates whether the execution outputs of the predicted SQL match the ground truth on the database. Few-shot learning was applied using 47 manually created examples, including questions, SQL queries, and corresponding answers. With CRP, GPT-4 achieves the highest performance (EM: 78.72\%, EX: 80.85\%), followed by GPT-4o Mini, Llama-3-8B-Instruct and GPT-3.5 Turbo, CRP consistently delivers better accuracy in most of LLMs, indicating the importance of detailed schema representation for SQL generation.

\renewcommand{\arraystretch}{0.9} 
\setlength{\tabcolsep}{4pt} 

\begin{table}[h!]
\centering
\small
\begin{tabularx}{\linewidth}{l *{4}{>{\centering\arraybackslash}X}}
\toprule
\multirow{2}{*}{\textbf{LLM}} & \multicolumn{2}{c}{\textbf{ODP}} & \multicolumn{2}{c}{\textbf{CRP}} \\
\cmidrule(lr){2-3} \cmidrule(lr){4-5}
 & \textbf{EM} & \textbf{EX} & \textbf{EM} & \textbf{EX} \\
\midrule
GPT-4           & 74.47\% & 78.72\% & 76.60\% & 80.85\% \\
GPT-4o Mini      & 76.60\% & 70.21\% & 78.72\% & 80.85\% \\
GPT-3.5 Turbo   & 38.30\% & 38.30\% & 25.53\% & 27.70\% \\
Llama-3-8B-Instruct         & 31.91\% & 29.79\% & 68.83\% & 46.81\% \\
\bottomrule
\end{tabularx}
\caption{SQL RAG results on the straightforward dataset.}
\label{tab:sql_rag}
\end{table}

Table \ref{tab:schema_prompt} presents the performance of Graph RAG, showing strong results across all models when using the schema prompt. GPT-4 leads with the highest accuracy (EM: 14.89\%, EX: 91.49\%), followed by GPT-4o Mini (EM: 10.64\%, EX: 89.36\%). 

The differing EM and EX results between SQL RAG and Graph RAG indicate the differences between the two methods. In SQL RAG, the data is highly structured, leading to more fixed SQL queries and higher EM scores whenever we execute it. In contrast, Graph RAG shows a much lower EM but high EX, indicating that the graph query language is more flexible and can generate different formats while still providing highly accurate answers.

\renewcommand{\arraystretch}{0.9} % 缩小行距
\setlength{\tabcolsep}{4pt} % 缩小列间距

\begin{table}[h!]
\centering
\small
\begin{tabular}{lcc}
\toprule
\multirow{2}{*}{\textbf{LLM}} & \multicolumn{2}{c}{\textbf{Schema Prompt}} \\
\cmidrule(lr){2-3}
 & \textbf{EM} & \textbf{EX} \\
\midrule
GPT-4           & 14.89\% & 91.49\% \\
GPT-4o Mini      & 10.64\% & 89.36\% \\
GPT-3.5 Turbo   & 10.64\% & 82.98\% \\

\bottomrule
\end{tabular}
\caption{Graph RAG Results with schema prompt on the straightforward dataset.}
\label{tab:schema_prompt}
\end{table}

\subsubsection{Specialized airport jargon, abbreviations, and incomplete questions}

As mentioned in the dataset section, we manually created a complicated, ambiguous dataset containing thousands of airport jargon, abbreviations, and incomplete questions. We classified these questions into six categories: Time Ambiguous Questions (TAQ), and Time With Ambiguous Flight Number Questions (TWAQ). Board Gate Questions (BGQ), Next Flight Questions (NFQ), Board Questions of Aircraft (BQA), and Ambiguous Flight Number Questions (AFQ).

Board Questions of Aircraft (BQA) and Ambiguous Flight Number Questions (AFQ) involve abbreviations and jargon, such as "Where is the delta?" and "At what gate is the 144?" Without the full airline names or additional flight details, these questions are challenging to answer. Time Ambiguous Questions (TAQ), Board Gate Questions (BGQ), and Time With Ambiguous Flight Number Questions (TWAQ) represent incomplete questions like "Which flight is currently at gate F09?" or "What's at C14?" These lack critical details such as flight numbers. Next Flight Questions (NFQ), on the other hand, are dynamic and will be discussed further in a later section.

We analyzed 220 questions in total to evaluate the robustness of the question classification prompt. Since large language models (LLMs) showed some variability in each time response, we employed a few-shot learning approach by integrating 60 carefully selected question classification examples within the context window into the prompt. These 60 examples included six different questions and their correct categories. We repeated the classification experiments five times on the same questions. The accuracies for these five times' classification runs were 90.45\%, 90.45\%, 90.91\%, 90.45\%, and 90.00\%, the average accuracy is 90.45\%. The low variance among these runs suggests our prompt is robust and effective. Few-shot learning with extensive examples significantly improved accuracy and ensured consistent performance for different question types.

\begin{figure}[t]
  \vspace{-5pt} 
  \includegraphics[width=0.9\columnwidth]{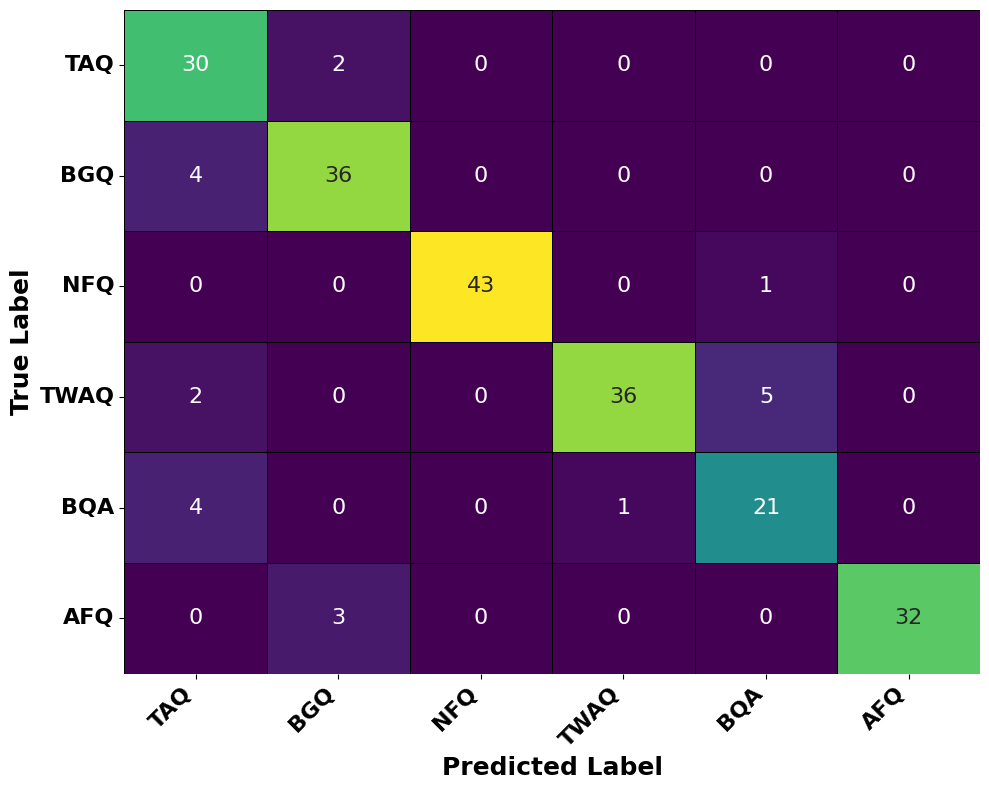}
  \vspace{-5pt} 
  \caption{Confusion Matrix of Question Classifications}
  \label{fig:confusion_matrix}
\end{figure}

The final classification results are shown in the Figure ~\ref{fig:confusion_matrix}.  Although most questions were classified correctly, about 22 questions were misclassified. However, TAQ and BGQ share the same subsequent step of extracting a gate number, so swapping them does not affect outcomes. Similarly, TWAQ and BQA both prompt users for additional information; hence the confusion between these two also does not have too much impact on final results. When TWAQ or BQA are misclassified as TAQ, the system fails to extract a gate number, returns ['0'], and prompts the user for more details before re-running RAG. Because subsequent steps rely on correct classification, we added additional measures to mitigate the impact of misclassification. Our experiments show that most errors occur within these similar categories, and we have worked to minimize them as much as possible. Further details on the question classification prompts are provided in Appendix~\ref{sec:appendix}.

\subsubsection{Dynamic questions}

The Next Flight Questions (NFQ) involves two situations: determining the next flight from the same airline or the same ramp. For the same airline, the answer is directly found in the table 'connecting flight number'. For the same ramp, we need to determine the expected on-ramp time for the current flight and then identify the closest expected on-ramp time for other flights at that ramp. Dynamic questions require additional calculation and reasoning. for example, if the question is 'What is the expected on-ramp time for the connecting flight of DL0123?'  we must first identify DL0123's next connecting flight, then we can find its expected on-ramp time. We created a dataset of 30 reasoning questions to test RAG methods. As shown in Table \ref{tab:rag_pipelines}, Graph RAG performed well, leveraging graph relationships for improved retrieval.

\begin{table}[h!]
\centering
\small
\begin{tabular}{lc}
\toprule
\textbf{RAG Pipeline}    & \textbf{Reasoning Question Dataset} \\
\midrule
Graph RAG       & 68.75\% \\
SQL RAG         & 6.25\% \\
Traditional RAG & 9.38\% \\
\bottomrule
\end{tabular}
\caption{Performance of different RAG pipelines on the reasoning question dataset}
\label{tab:rag_pipelines}
\end{table}

\subsection{RQ2: How to reduce hallucinations?}

Hallucinations mainly happen in traditional RAG when LLMs generate flight destinations not included in our dataset. This issue mainly exists in responses to complex and ambiguous queries. After performing question classification and retrieving flight information, we conducted few-shot learning with 20 examples, observing a hallucination rate of approximately 10\%. This phenomenon is likely due to the excessive amount of information included in the input prompts for traditional RAG, which increases the likelihood of hallucination compared to SQL RAG and Graph RAG. Additionally, airline companies often reuse flight numbers, leading to conflicting data in LLM training and causing the generation of information absent from the dataset.

SQL RAG and Graph RAG reduce hallucinations by converting natural language questions into SQL or Cypher queries. Thereby, the input to the LLM is accurate data, which significantly reduces hallucinations. However, if the question requires a lot of context, the conversion to a query may fail.

It is important to note that hallucinations are not common even in traditional RAG and are not eliminated in SQL RAG or Graph RAG. Additionally, calculating the exact accuracy or rate of hallucinations across these RAG methods is challenging. However, SQL RAG and Graph RAG tend to reduce the occurrence of hallucinations compared to traditional RAG. Given the high safety requirements in airport and aviation environments, SQL RAG and Graph RAG are safer for aviation operations. Both support dynamic storage of real-time flight information. Among them, Graph RAG performs better due to its stronger reasoning capabilities, enabling it to handle more complex queries effectively. More details of the experiment are provided in the Appendix~\ref{sec:appendix}.

\section{Conclusion}
\label{sec:conclusion}

Our evaluation of three RAG methods shows that of the traditional RAG methods, BM25+GPT-4 is more efficient than other methods, because of the terminology used in the airport. However, traditional RAG can produce hallucinations, which poses a safety risk. SQL RAG and Graph RAG produce fewer hallucinations, and Graph RAG on top has higher accuracy.
Our overall system effectively handles specialized airport terminology through question classification and prompt engineering; specifically, we address airport jargon and abbreviations. Graph RAG is particularly effective in handling reasoning tasks and questions about dynamic data, making it efficient in the airport domain.

\section{Future Work}

In our current research, the experiments are based on a static environment that does not capture any real-time changes such as delays or gate changes. In future research, we plan to connect the system with live APIs that provide real-time flight status and gate information, so that the system can dynamically retrieve and use real-time data. Another limitation is the relatively small size of our current dataset. In future work, we want to significantly expand and diversify the dataset. A larger and more diverse dataset will help ensure that our performance improvements hold across different scenarios and strengthen the validity of our conclusions.

\section*{Limitations}

We openly acknowledge that this study is not a finalized product but an initial research investigation. The system's performance in the real world has not been demonstrated; it is a prototype that was tested in a controlled environment. Moreover, our evaluation is specific to the Schiphol airport domain; adapting the model to other airports or domains may present new challenges. Any deployment would need careful incremental trials, user feedback, and regulatory compliance checks to meet the high-reliability standards expected in aviation contexts.

\section*{Ethics Policy}

Our research uses a dataset that has been authorized by Amsterdam Airport Schiphol and contains outdated flight information. Most of the flight information is publicly available online and does not include sensitive information. In this paper, the dataset is not publicly released, and it is only used to discuss its structure and to provide examples of question-answer pairs. No personal or confidential data are involved. Importantly, this work is an exploratory study focused on benchmarking performance in a controlled environment without impacting actual airport operations. We have implemented methods to reduce AI hallucinations—a key safety concern in this domain. However, any future deployment would require additional security reviews and strong safeguards to prevent misuse.

\section*{Acknowledgments}
This work was supported in part by the Deutsche Forschungsgemeinschaft (DFG, German Research Foundation)  under Research Unit FOR 5339 (Project No.459291153). We also gratefully acknowledge Amsterdam Airport Schiphol for providing the raw dataset used in this study, and we thank our co-authors from the Royal Schiphol Group for their valuable contributions.

\bibliography{custom}

\begin{thebibliography}{21}
\providecommand{\natexlab}[1]{#1}

\bibitem[{Barrasa et~al.(2023)Barrasa, Webber, and
  Webber}]{barrasa2023building}
J.~Barrasa, J.~Webber, and J.~Webber. 2023.
\newblock \href {https://books.google.de/books?id=Ztb5zgEACAAJ} {\emph{Building
  Knowledge Graphs: A Practitioner's Guide}}.
\newblock O'Reilly.

\bibitem[{Chen and Wiseman(2023)}]{chen2023bm25queryaugmentationlearned}
Xiaoyin Chen and Sam Wiseman. 2023.
\newblock \href {https://arxiv.org/abs/2305.14087} {Bm25 query augmentation
  learned end-to-end}.
\newblock \emph{Preprint}, arXiv:2305.14087.

\bibitem[{Edge et~al.(2024)Edge, Trinh, Cheng, Bradley, Chao, Mody, Truitt, and
  Larson}]{edge2024localglobalgraphrag}
Darren Edge, Ha~Trinh, Newman Cheng, Joshua Bradley, Alex Chao, Apurva Mody,
  Steven Truitt, and Jonathan Larson. 2024.
\newblock \href {https://arxiv.org/abs/2404.16130} {From local to global: A
  graph rag approach to query-focused summarization}.
\newblock \emph{Preprint}, arXiv:2404.16130.

\bibitem[{Gao et~al.(2023)Gao, Wang, Li, Sun, Qian, Ding, and
  Zhou}]{gao2023texttosqlempoweredlargelanguage}
Dawei Gao, Haibin Wang, Yaliang Li, Xiuyu Sun, Yichen Qian, Bolin Ding, and
  Jingren Zhou. 2023.
\newblock \href {https://arxiv.org/abs/2308.15363} {Text-to-sql empowered by
  large language models: A benchmark evaluation}.
\newblock \emph{Preprint}, arXiv:2308.15363.

\bibitem[{Gao et~al.(2024)Gao, Xiong, Gao, Jia, Pan, Bi, Dai, Sun, Wang, and
  Wang}]{gao2024retrievalaugmented}
Yunfan Gao, Yun Xiong, Xinyu Gao, Kangxiang Jia, Jinliu Pan, Yuxi Bi, Yi~Dai,
  Jiawei Sun, Meng Wang, and Haofen Wang. 2024.
\newblock \href {https://arxiv.org/abs/2312.10997} {Retrieval-augmented
  generation for large language models: A survey}.
\newblock \emph{Preprint}, arXiv:2312.10997.

\bibitem[{George and Rajan(2022)}]{10039758}
Godwin George and Rajeev Rajan. 2022.
\newblock \href {https://doi.org/10.1109/INDICON56171.2022.10039758} {A
  faiss-based search for story generation}.
\newblock In \emph{2022 IEEE 19th India Council International Conference
  (INDICON)}, pages 1--6.

\bibitem[{Guo et~al.(2023)Guo, Tian, Tang, Li, Wen, Wang, and
  Wang}]{guo2023retrievalaugmentedgpt35basedtexttosqlframework}
Chunxi Guo, Zhiliang Tian, Jintao Tang, Shasha Li, Zhihua Wen, Kaixuan Wang,
  and Ting Wang. 2023.
\newblock \href {https://arxiv.org/abs/2307.05074} {Retrieval-augmented
  gpt-3.5-based text-to-sql framework with sample-aware prompting and dynamic
  revision chain}.
\newblock \emph{Preprint}, arXiv:2307.05074.

\bibitem[{Jacobs and Jaschke(2024)}]{jacobs2024leveraging}
Sven Jacobs and Steffen Jaschke. 2024.
\newblock \href {https://arxiv.org/abs/2405.06681} {Leveraging lecture content
  for improved feedback: Explorations with gpt-4 and retrieval augmented
  generation}.
\newblock \emph{Preprint}, arXiv:2405.06681.

\bibitem[{Jafari et~al.(2021)Jafari, Maurya, Nagarkar, Islam, and
  Crushev}]{jafari2021surveylocalitysensitivehashing}
Omid Jafari, Preeti Maurya, Parth Nagarkar, Khandker~Mushfiqul Islam, and
  Chidambaram Crushev. 2021.
\newblock \href {https://arxiv.org/abs/2102.08942} {A survey on locality
  sensitive hashing algorithms and their applications}.
\newblock \emph{Preprint}, arXiv:2102.08942.

\bibitem[{Jegou et~al.(2017)Jegou, Douze, and Johnson}]{faiss2017}
Hervé Jegou, Matthijs Douze, and Jeff Johnson. 2017.
\newblock \href
  {https://engineering.fb.com/2017/03/29/data-infrastructure/faiss-a-library-for-efficient-similarity-search/}
  {Faiss: A library for efficient similarity search}.
\newblock Facebook AI Research, Data Infrastructure, ML Applications.
\newblock Accessed: insert access date.

\bibitem[{Lewis et~al.(2021)Lewis, Perez, Piktus, Petroni, Karpukhin, Goyal,
  Küttler, Lewis, tau Yih, Rocktäschel, Riedel, and
  Kiela}]{lewis2021retrievalaugmentedgenerationknowledgeintensivenlp}
Patrick Lewis, Ethan Perez, Aleksandra Piktus, Fabio Petroni, Vladimir
  Karpukhin, Naman Goyal, Heinrich Küttler, Mike Lewis, Wen tau Yih, Tim
  Rocktäschel, Sebastian Riedel, and Douwe Kiela. 2021.
\newblock \href {https://arxiv.org/abs/2005.11401} {Retrieval-augmented
  generation for knowledge-intensive nlp tasks}.
\newblock \emph{Preprint}, arXiv:2005.11401.

\bibitem[{Liu et~al.(2018)Liu, Sheng, Wei, and Yang}]{8492945}
Cai-zhi Liu, Yan-xiu Sheng, Zhi-qiang Wei, and Yong-Quan Yang. 2018.
\newblock \href {https://doi.org/10.1109/IRCE.2018.8492945} {Research of text
  classification based on improved tf-idf algorithm}.
\newblock In \emph{2018 IEEE International Conference of Intelligent Robotic
  and Control Engineering (IRCE)}, pages 218--222.

\bibitem[{Louis et~al.(2023)Louis, van Dijck, and
  Spanakis}]{louis2023interpretable}
Antoine Louis, Gijs van Dijck, and Gerasimos Spanakis. 2023.
\newblock \href {https://doi.org/10.48550/arXiv.2309.17050} {Interpretable
  long-form legal question answering with retrieval-augmented large language
  models}.
\newblock In \emph{Proceedings of the Thirty-Eighth AAAI Conference on
  Artificial Intelligence (AAAI-24)}, Maastricht University, Law \& Tech Lab.
\newblock Under review. Code available at https://arxiv.org/abs/2309.17050.

\bibitem[{Mao et~al.(2020)Mao, Qu, Xie, Ren, and
  Han}]{mao2020multidocumentsummarizationmaximalmarginal}
Yuning Mao, Yanru Qu, Yiqing Xie, Xiang Ren, and Jiawei Han. 2020.
\newblock \href {https://arxiv.org/abs/2010.00117} {Multi-document
  summarization with maximal marginal relevance-guided reinforcement learning}.
\newblock \emph{Preprint}, arXiv:2010.00117.

\bibitem[{Mikolov et~al.(2013)Mikolov, Chen, Corrado, and
  Dean}]{mikolov2013efficientestimationwordrepresentations}
Tomas Mikolov, Kai Chen, Greg Corrado, and Jeffrey Dean. 2013.
\newblock \href {https://arxiv.org/abs/1301.3781} {Efficient estimation of word
  representations in vector space}.
\newblock \emph{Preprint}, arXiv:1301.3781.

\bibitem[{Rajkumar et~al.(2022)Rajkumar, Li, and
  Bahdanau}]{rajkumar2022evaluatingtexttosqlcapabilitieslarge}
Nitarshan Rajkumar, Raymond Li, and Dzmitry Bahdanau. 2022.
\newblock \href {https://arxiv.org/abs/2204.00498} {Evaluating the text-to-sql
  capabilities of large language models}.
\newblock \emph{Preprint}, arXiv:2204.00498.

\bibitem[{Robert~Lee(2024)}]{microsoft_azure}
Heidi~Steen Robert~Lee. 2024.
\newblock Hybrid search using vectors and full text in azure ai search.
\newblock
  \emph{https://learn.microsoft.com/en-us/azure/search/hybrid-search-overview}.

\bibitem[{Robertson(2004)}]{article}
Stephen Robertson. 2004.
\newblock \href {https://doi.org/10.1108/00220410410560582} {Understanding
  inverse document frequency: On theoretical arguments for idf}.
\newblock \emph{Journal of Documentation - J DOC}, 60:503--520.

\bibitem[{Robertson and Zaragoza(2009)}]{article2}
Stephen Robertson and Hugo Zaragoza. 2009.
\newblock \href {https://doi.org/10.1561/1500000019} {The probabilistic
  relevance framework: Bm25 and beyond}.
\newblock \emph{Foundations and Trends in Information Retrieval}, 3:333--389.

\bibitem[{Sarmah et~al.(2024)Sarmah, Hall, Rao, Patel, Pasquali, and
  Mehta}]{sarmah2024hybridragintegratingknowledgegraphs}
Bhaskarjit Sarmah, Benika Hall, Rohan Rao, Sunil Patel, Stefano Pasquali, and
  Dhagash Mehta. 2024.
\newblock \href {https://arxiv.org/abs/2408.04948} {Hybridrag: Integrating
  knowledge graphs and vector retrieval augmented generation for efficient
  information extraction}.
\newblock \emph{Preprint}, arXiv:2408.04948.

\bibitem[{Xu et~al.(2024)Xu, Jain, and
  Kankanhalli}]{xu2024hallucinationinevitableinnatelimitation}
Ziwei Xu, Sanjay Jain, and Mohan Kankanhalli. 2024.
\newblock \href {https://arxiv.org/abs/2401.11817} {Hallucination is
  inevitable: An innate limitation of large language models}.
\newblock \emph{Preprint}, arXiv:2401.11817.

\end{thebibliography}

\appendix

\section{Appendix}
\label{sec:appendix}

\subsection{Data Generation}
In this section, we explained the methods to generate ground truth datasets including Question generation and Question classification. 

\subsubsection{Question Classification}

To classify the questions, a flight information dataset is used to create different categories of questions. The flight information dataset contains information for thousands of flights which include several key items: The flight number identifies a specific flight, aircraft category, bus gate, bus service needed (remote or none), flight UID, direction (departure or arrival), ramp, main ground handler, expected on-ramp time, expected off-ramp time, connecting flight number, connecting flight UID, modified date and time, previous ramp, aircraft registration, flight state, MTT (minimum transfer time), MTT single leg, EU indicator, safe town airport (J or P), scheduled block time, best block time, expected block time, expected tow-in time, expected tow-off time, actual final approach time, actual block time, actual take-off time, actual boarding time, actual tow-in request time, actual tow-off time, actual on-ramp time, actual off-ramp time, flight nature, push back, and pier.  Based on this flight information, we make some classifications for the questions. 

\begin{figure}[h!]
    \centering
    \includegraphics[scale=0.4]{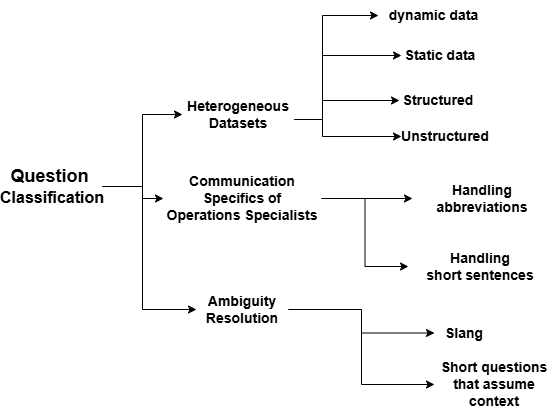}
    \caption{Question Classifications}
    \label{fig:1}
\end{figure}

The Question Classification pipeline is shown in Figure \ref{fig:1} Multiple types of questions need to be addressed in the project. Firstly, there are Heterogeneous datasets, which contain different formats of datasets, including static data and dynamic data. Static data are flight information that remains constant for example, flight number, flight uid, EU indicator, flight nature, etc. while dynamic data are the flight information that changes dynamically, such as the time information expected on-ramp time, expected off-ramp time, modified date and time, connecting flight number, etc. this information are changed dynamically. Secondly, there are communication specifics of operations specialists' questions, which require handling abbreviations and short sentences. Thirdly, there are ambiguity resolution questions, which include ambiguity questions such as airport slang, and short questions that assume context. For example, some user questions are very short and not clear, such as "What is at A74?" or "Delta airline, any information?" These types of questions are also taken into consideration.

To evaluate how effectively different retrieval methods performed, several tests were run. Two ground truth datasets were created: a straightforward dataset and a complicated and ambiguous dataset. The straightforward dataset contains questions without any ambiguities, which can be directly retrieved from flight information articles, Examples of straightforward questions were: "What category of aircraft is designated for flight KL1000?", "Which ramp is assigned for flight KL1000?", and "What time is the expected on-ramp for flight KL0923?" Such questions are easily identified for retrieval methods, which enables the selection of the most relevant articles. 

The second type of dataset is a complicated and ambiguous dataset; it is made of variables that may be ambiguous and missing from the flight information dataset. Examples of such questions are: "It is now 2023-05-14 18:07:34+0000. Which flight is at gate B24?" or "Can you tell me which flight is scheduled at gate B24 for 2023-05-14 18:07:34+0000?" The gate number remains a constant variable in this case, but the given time indicates a random variable that is one hour before the scheduled block time in the flight table. When there aren't sufficient keywords, the pipeline finds it very challenging to find the exact correct content to generate correct answers. Besides, the complicated and ambiguous dataset also includes questions with ambiguous information, such as "Which flight is at gate B24?" These questions lack specific time and aircraft. Which results in multiple flight information that mention gate B24. In addition, many articles contain the B24 gate, in this case, BM25 is capable of retrieving all articles containing B24 as a keyword, and it returns correct articles within the top 30 results, indicating that the relevant article is among these top 30 articles.

\subsubsection{Question generation}

This part includes how to generate benchmark datasets. 

As previously mentioned, we manually created two benchmark datasets: a straightforward dataset and a complicated/ambiguous dataset. The straightforward dataset contains questions that can be directly answered using flight information tables. In contrast, the complicated/ambiguous dataset includes more vague questions that depend on variables like time, airline, and flight number. For example, the question "Which flight is in B24?" could refer to many flights, so additional information is needed for an accurate answer. To generate the straightforward dataset, we created question templates with placeholders like: "Is there a problem with aircraft separation at  $<$gate\_nr$>$?"
"What airlines have flights departing from gate  $<$gate\_nr$>$?"; "Can you tell me the aircraft category for flight $<$flight\_number$>$?" We then manually filled these placeholders with actual gate and flight numbers from the flight information table.

% as we mentioned in the dataset part, we manually created two benchmark datasets, a straightforward dataset, and a complicated and ambiguous dataset. Straightforward questions are used for some of the questions that can be found directly in flight information tables. However, complicated/ambiguous question datasets include more ambiguous questions that will be changed based on different times, airlines, and flight numbers. For example "Which flight in B24" This question will retrieve lots of relevant flights, so more information is needed to get the correct answer. When we generate a straightforward question dataset, we generate some straightforward questions template with the place holder like: Is there a problem with aircraft separation at $<$gate\_nr$>$ ?", " What airlines have flights departing from gate $<$gate\_nr$>$ ?", "Can you tell me the aircraft category for flight $<$flight\_number$>$", etc. then the exact values in the placeholder are queried manually from the flight information table, for example, we can fill in the exact flight number, and gate number from the flight information table to the questions that is generated.

\begin{figure}[h!]
    \centering
    \includegraphics[width=0.45\textwidth]{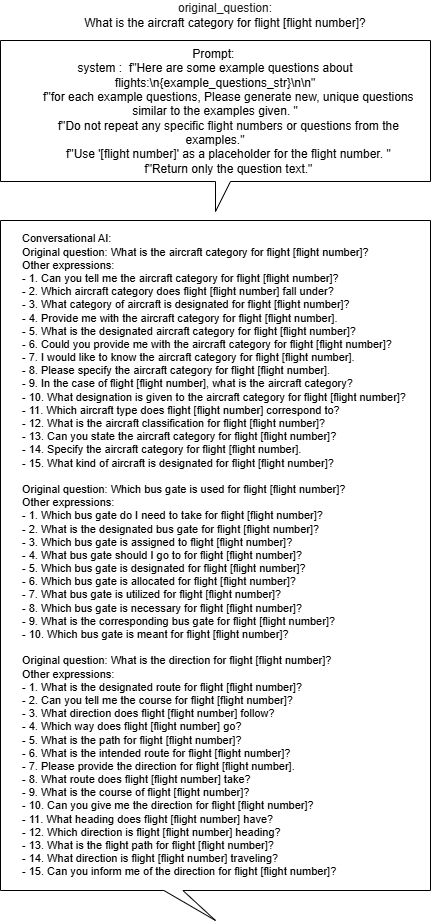}
    \caption{Dataset generation examples of straightforward questions}
    \label{fig:straightforward_generation}
\end{figure}

To enrich our questions, we used language models to generate more variations. To enrich our questions, we used language models to generate more variations. For example, as Figure \ref{fig:straightforward_generation} shows, we took the question "What is the aircraft category for flight [flight\_number]?" and prompted the model: We provide prompts like: "For each example question, please generate new, unique questions similar to the examples given, Do not repeat any specific flight numbers or questions from the examples. Use '[flight number]' as a placeholder for the flight number. Return only the question text." The same method will also be used for other types of straightforward questions. After that, the exact values in the placeholders such as [flight number], [ramp], [bus gate], etc., will be queried from the flight information dataset manually.Using this approach, we generated thousands of straightforward questions to test the performance of the conversational AI system. During our experiments, we randomly selected 150-200 question-answer pairs from the straightforward dataset. When evaluating different RAG methods and the performance of language models, we manually labeled each response as 'True' or 'False' to calculate accuracy.

\begin{figure}[h!]
    \centering
    \includegraphics[width=0.45\textwidth]{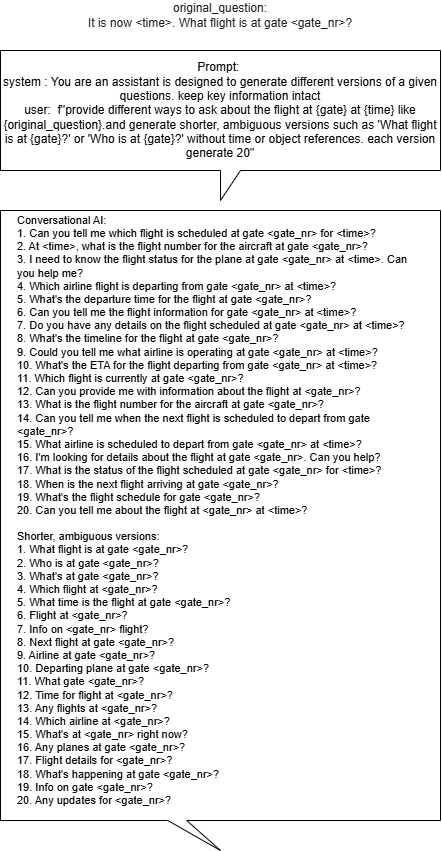}
    \caption{Dataset generation examples of complicated/ambiguous questions}
    \label{fig:complicated_generation}
\end{figure}

Similarly, for the ambiguous and complicated dataset, examples of such questions include: "It is now 2023-05-14 18:07:34+0000. Which flight is at gate B24?" or "Can you tell me which flight is scheduled at gate B24 for 2023-05-14 18:07:34+0000?" In these cases, the gate number is constant, but the provided time varies—usually set to one hour before the scheduled block time in the flight table. When keywords are insufficient, the system struggles to find the exact information needed for correct answers. The dataset also contains questions with ambiguous information, such as: "Which flight is at gate B24?" These questions lack specific time or aircraft details, leading to multiple flights associated with gate B24. As Figure \ref{fig:complicated_generation} showed, during our experiments, we classified these complicated questions. We randomly selected 100-200 question-answer pairs, manually labeled their categories for question classification, and marked their prompt engineering results as 'True' or 'False' after classification.

\subsection{Experiment}
\subsubsection{Question Classification and Prompt Engineering}

\begin{figure*}[h!]
    \centering
    \includegraphics[width=0.8\textwidth]{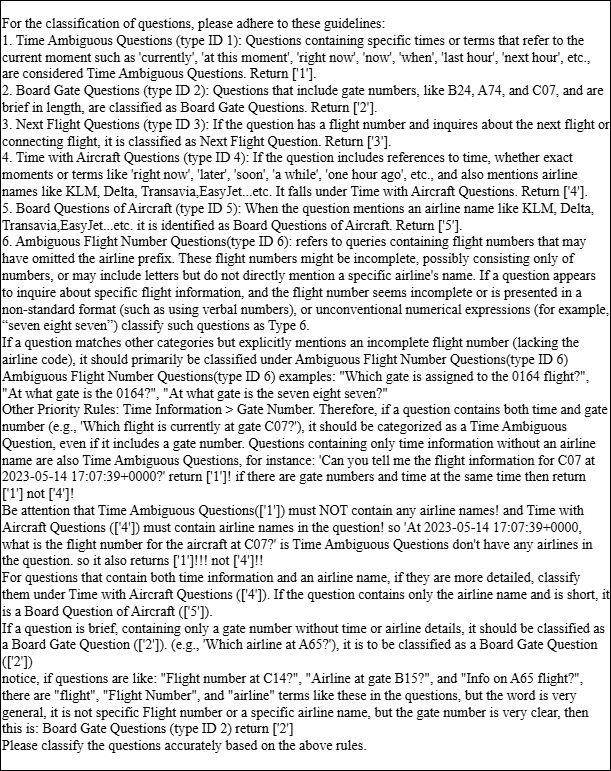}
    \caption{Prompts of Question Classifications}
    \label{fig:QA_class_prompt}
\end{figure*}

In this step, questions are categorized into six types: Time Ambiguous Questions, Board Gate Questions, Next Flight Questions, Time with Airline Questions, Board Questions of Airline, and Ambiguous Flight Number Questions. The definitions of these question types are given to LLMs, and prompt them to response values from ['1'] to ['6']. Time Ambiguous Questions are questions that include specific times or terms referring to the current moment such as 'currently', 'at this moment', 'right now', 'now', 'when', 'last hour', 'next hour', etc., returning ['1']. Board Gate Questions are the questions that include gate numbers, like B24, A74, and C07, and are brief in length. returning ['2']. Next Flight Questions are the questions that inquire about a flight number and its next or connecting flight. returning ['3']. Time with Aircraft Questions involve references to time---exact moments or terms like 'right now', 'later', 'soon', 'a while', 'one hour ago', etc., and also mentions airline names like KLM, Delta, Transavia, EasyJet...etc, returning ['4']. Board Questions of Aircraft includes airline names, such as KLM, Delta, Transavia, EasyJet, etc.. returning ['5']. Ambiguous Flight Number Questions are the queries containing flight numbers that may have been ignored in the airline prefix, for example, "Which gate is assigned to the 0164 flight?", "At what gate is the 0164?". These flight numbers might be incomplete, possibly consisting only of numbers, or may include letters but do not directly mention a specific airline's name, then these are Ambiguous Flight Number Questions return ['6']. The details of this prompt are shown in the Figure \ref{fig:QA_class_prompt}

After classifying questions, we use different prompts for each type. For Time Ambiguous Questions and Board Gate Questions, we direct them to prompts that extract gate numbers from the queries. In traditional Retrieval Augmented Generation (RAG), we can query the ramp or gate table for language models. For example, the ambiguous question "Which airline at A74?" is reformatted to extract the ramp number like ['ramp': 'A74']. If it does not extract the ramp information successfully, it will return ['0']. After testing these prompts, we achieve over 80\% accuracy in extracting gate numbers. We then retrieve all tables containing these gate numbers from the flight information database. These tables, along with the original question, are used to generate answers. SQL RAG and Graph RAG directly generate query languages based on the ramp and gate numbers.

Questions that include airline names, like Time with Aircraft Questions and Board Questions of Aircraft, don't provide enough information because the airport has many flights from the same airline. Therefore, we prompt users to provide more details. For example, if someone asks, "Which flight is at Delta?", we respond: "I cannot determine the specific location of the Delta flight with the information provided. Please provide additional information like: - Flight UID (Unique Identifier) - Flight Number (Flight\_NR) - Aircraft Registration - Connecting Flight UID (The UID of any connected flight provided by the airline) - Connecting Flight Number (The number of any connected flight provided by the airline). If you do not have this information, I can still attempt to process your query but it might require additional search time. In this case, please let me know if you are looking for information about the Ramp (Gate), Bus Gate, or Pier." After the user provides more information, we use the RAG methods again.

For Next Flight Questions, there are two scenarios: the next flight is from the same airline or the same departure ramp. If it's the same airline, we return the connecting\_flight\_nr from the table. If it's the same ramp, we find all flights at that ramp and identify the one with the closest on-ramp time. To handle these questions, we write prompts for the RAG methods to find the relevant results. For the Ambiguous Flight Number Questions, we would like to extract the number and match it with the real-time flight APIs to find the relevant flights that contain those mentioned numbers. However, Since we were researching in the static environments, we responded: " We could not find more information about the flight number you mentioned, could you please provide us with more information?"

\subsubsection{SQL RAG}
\begin{figure}[h!]
    \centering
    \includegraphics[width=0.45\textwidth]{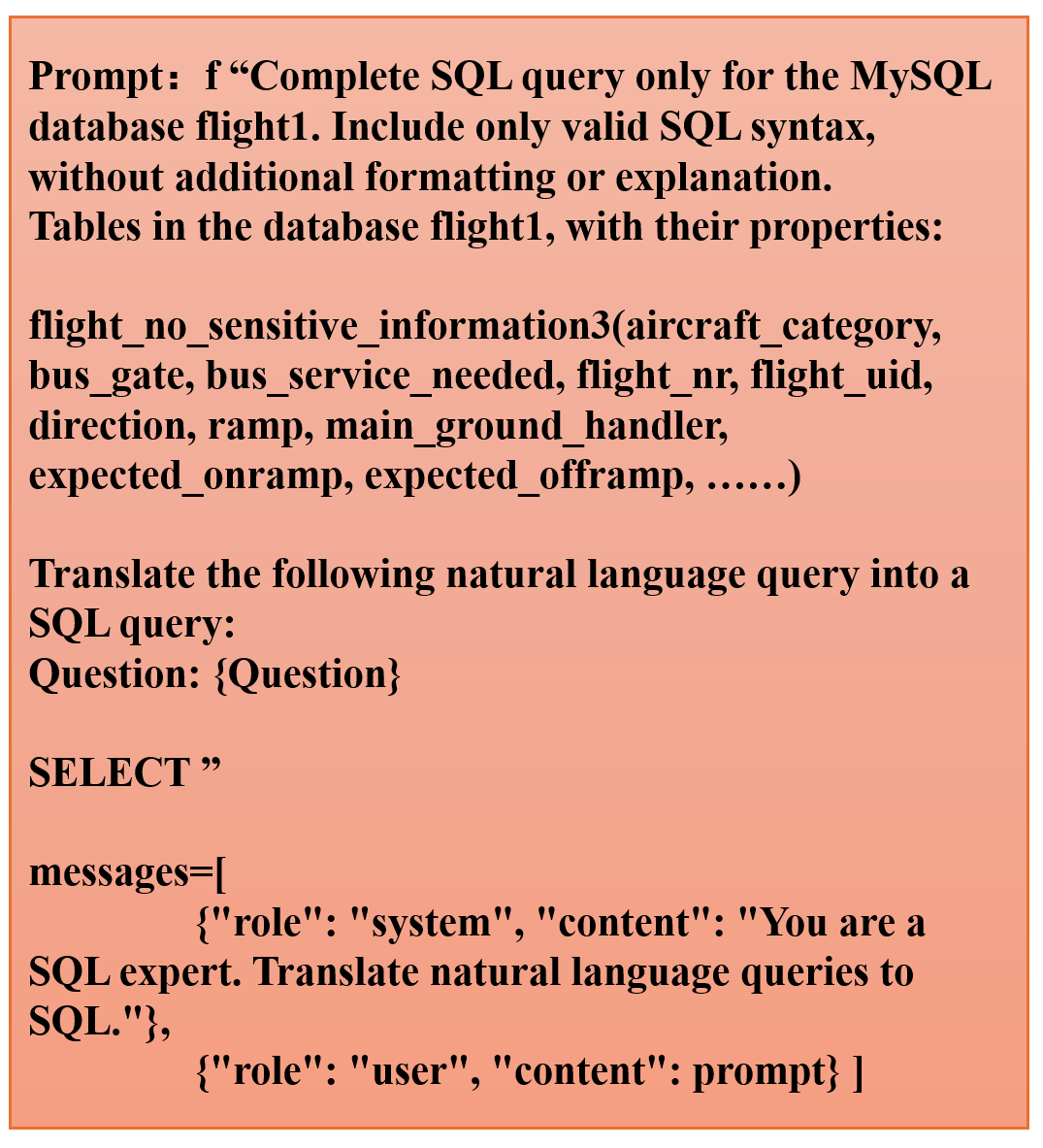}
    \caption{ODP Prompt for SQL RAG}
    \label{fig:sql_rag_odp}
\end{figure}

\begin{figure}[h!]
    \centering
    \includegraphics[width=0.45\textwidth]{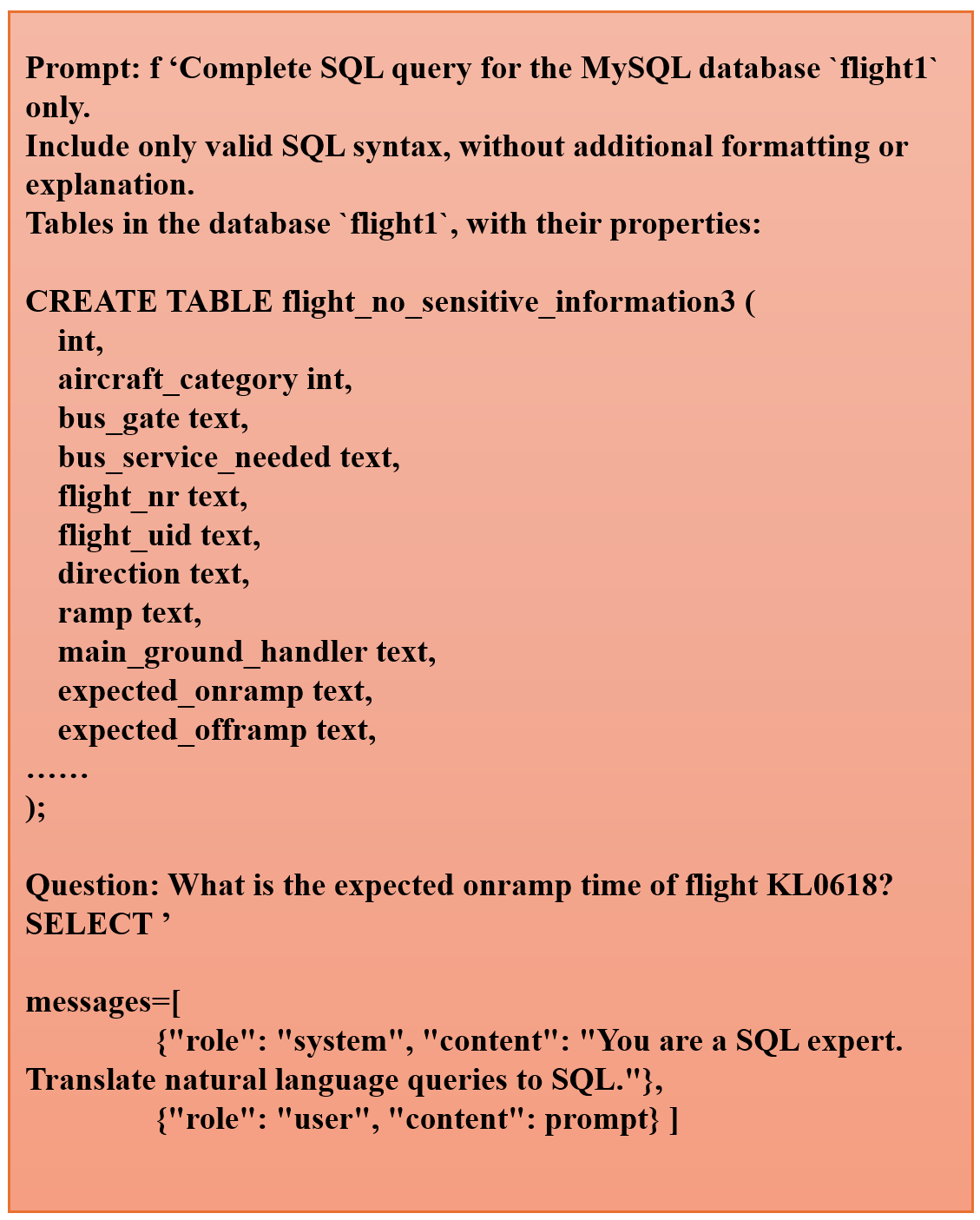}
    \caption{CRP Prompt for SQL RAG}
    \label{fig:sql_rag_crp}
\end{figure}

the OpenAI Demonstration Prompt(ODP) and Code Representation Prompt(CRP) prompts are showed in Figure \ref{fig:sql_rag_odp}, Figure \ref{fig:sql_rag_crp}

The ODP, as shown in Figure \ref{fig:sql_rag_odp}, focuses on simplicity and explicit rules. It lists table names and their respective columns without additional data types or constraints. The ODP style emphasizes straightforward task instructions, such as “Include only valid SQL syntax, without additional formatting or explanation” guiding the model to generate SQL queries directly without unnecessary explanations.

In contrast, the CRP, shown in Figure \ref{fig:sql_rag_crp}, adopts a detailed SQL-style schema description. This approach uses CREATE TABLE statements to include comprehensive database information, such as column types and relationships (e.g., primary and foreign keys). By simulating database creation scripts, CRP uses the model's coding capabilities to enhance query precision, especially for complex databases with intricate relationships.

ODP is suitable for simpler, direct tasks, while CRP is better for handling more complex databases with comprehensive schema context.

\subsubsection{Graph RAG in dynamic dataset}
\begin{figure*}[h!]
    \centering
    \includegraphics[width=0.5\textwidth]{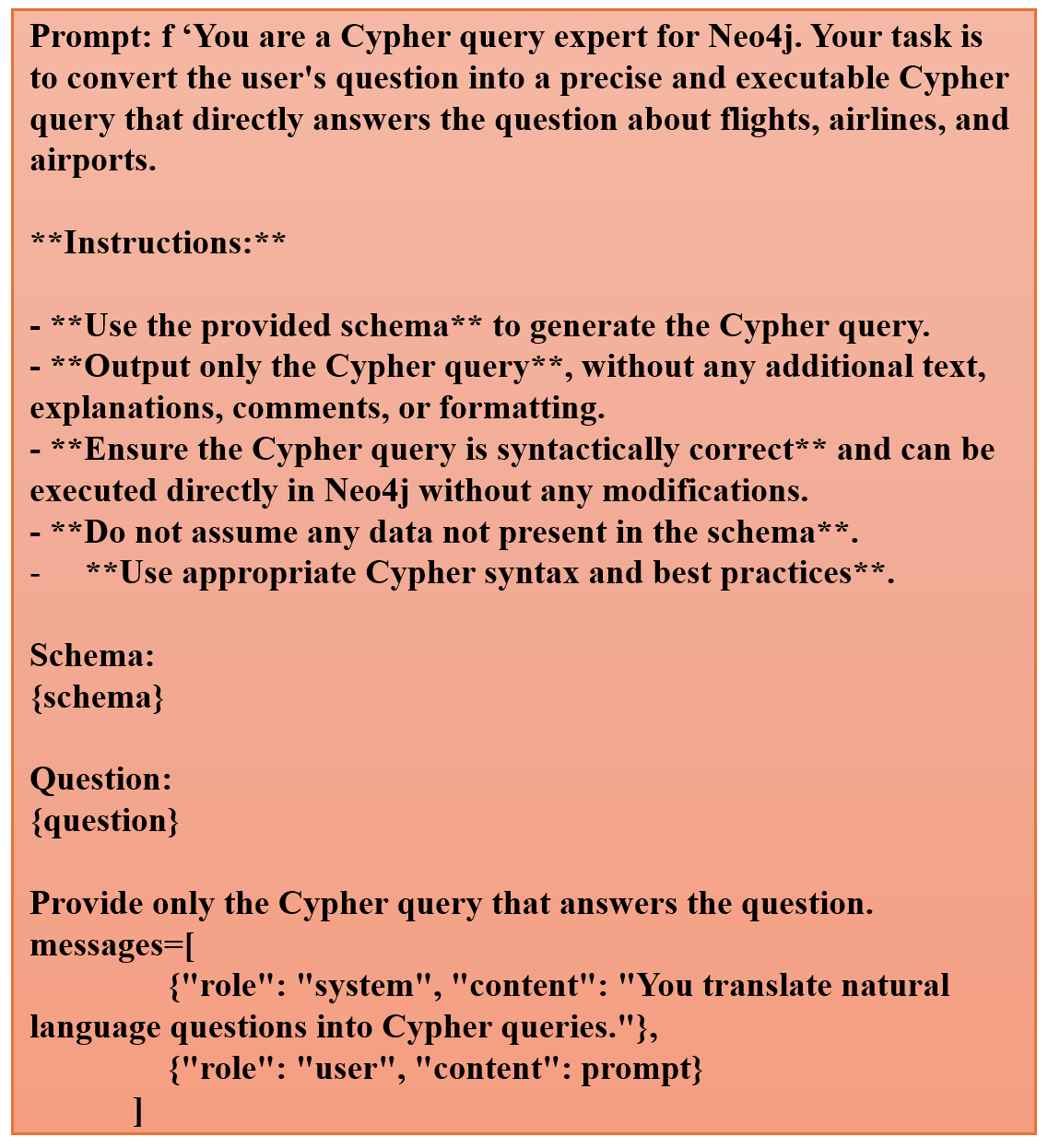}
    \caption{Graph RAG Prompt}
    \label{fig:graph_rag_prompt}
\end{figure*}

\begin{figure*}[h!]
    \centering
    \includegraphics[width=1\textwidth]{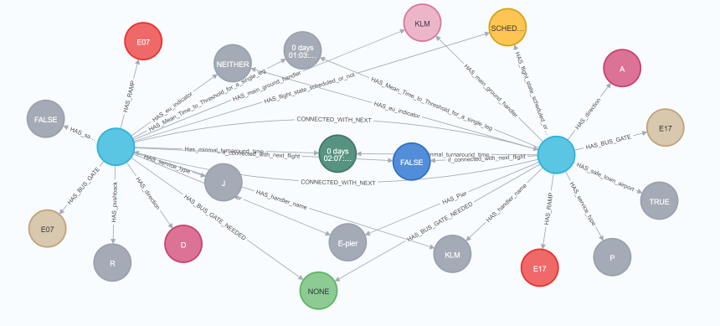}
    \caption{Graph RAG in Reasoning Questions}
    \label{fig:graph_rag_reasoning}
\end{figure*}

The Prompt of Graph RAG is shown in Figure \ref{fig:graph_rag_prompt}, focusing on guidelines for writing a Cypher query. The schema is extracted from the Neo4j graph database using the APOC plugin, specifically through 'CALL apoc.meta.schema() YIELD value RETURN value', and then used in the prompts. As shown in Figure \ref{fig:graph_rag_reasoning}, Graph RAG enables flights to be connected through their relationships, allowing retrieval of detailed information about connecting flights. In contrast, traditional RAG and SQL RAG treat connecting flights as merely a column in the table, limiting access to further relational information.

\subsubsection{Hallucination Analyses}

This section provides additional information about hallucinations in different RAG methods.

\begin{figure*}[h!]
    \centering
    \includegraphics[width=0.8\textwidth]{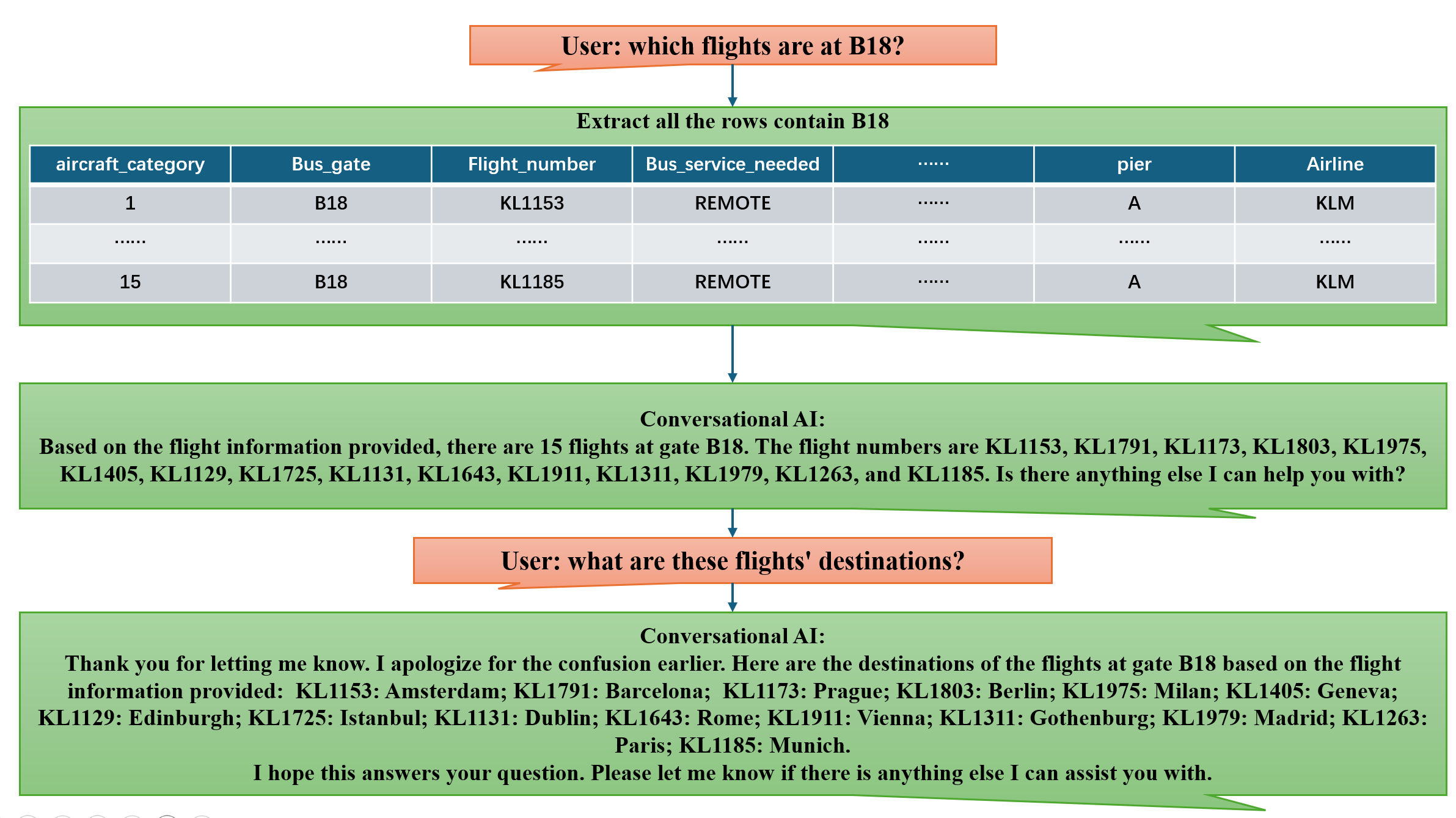}
    \caption{Hallucination case for traditional RAG}
    \label{fig:halllucination_tradition}
\end{figure*}

The hallucination in traditional RAG is illustrated in Figure \ref{fig:halllucination_tradition}. Using the question "Which flights are at B18" as an example, this query is classified as a Board Gate Question (BGQ) during the question classification step. For traditional RAG, the gate number "B18" is extracted from the question, and all table rows containing "B18" are retrieved. These rows, along with the question itself, are then passed to the LLM to generate the final answer. However, due to the large amount of flight information stored in LLMs, hallucinations are more likely to happen if the retrieval process brings in too much unrelated information.

\begin{figure*}[h!]
    \centering
    \includegraphics[width=0.8\textwidth]{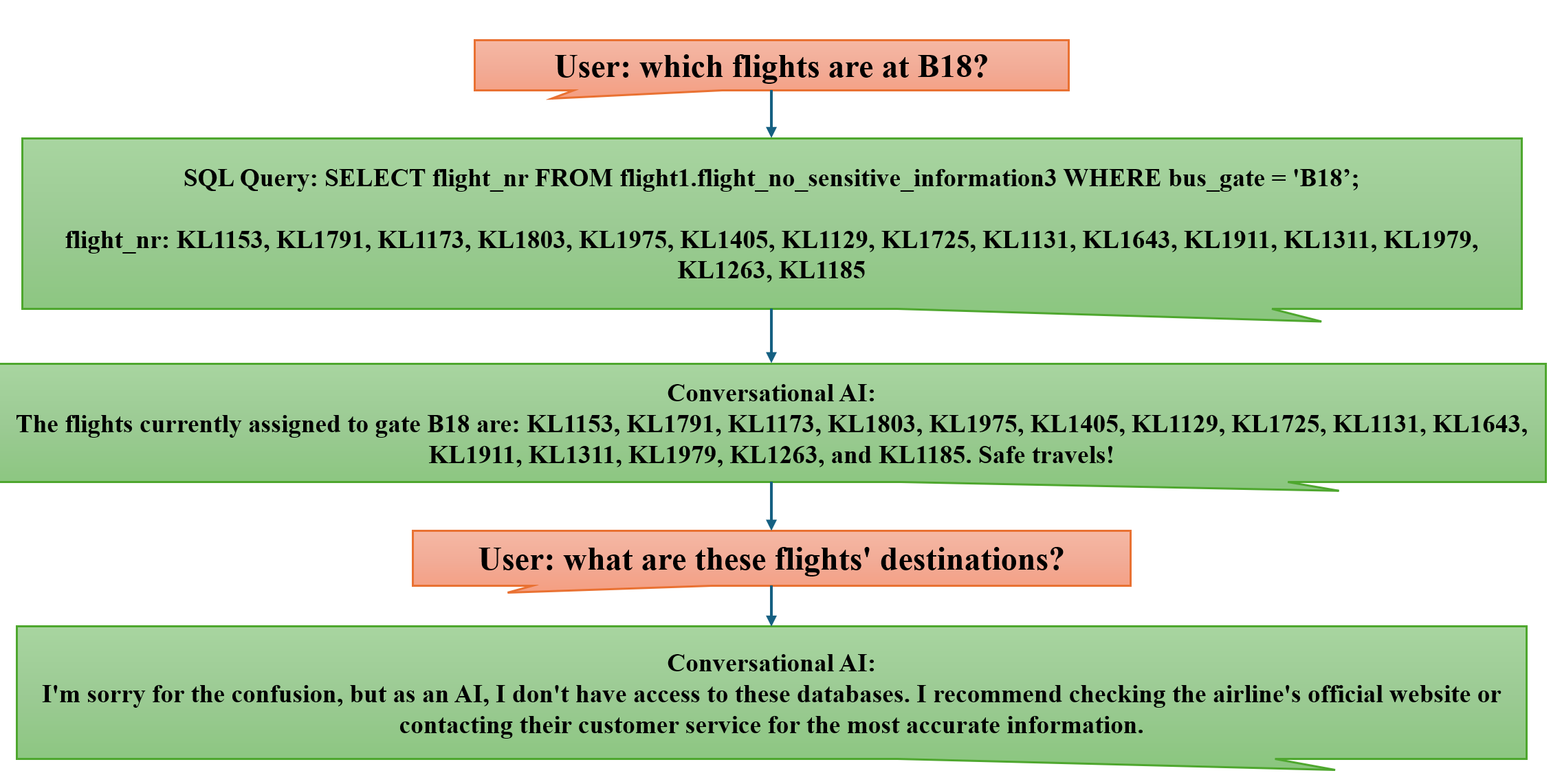}
    \caption{The same case for SQL RAG}
    \label{fig:halllucination_sql}
\end{figure*}

\begin{figure*}[h!]
    \centering
    \includegraphics[width=0.8\textwidth]{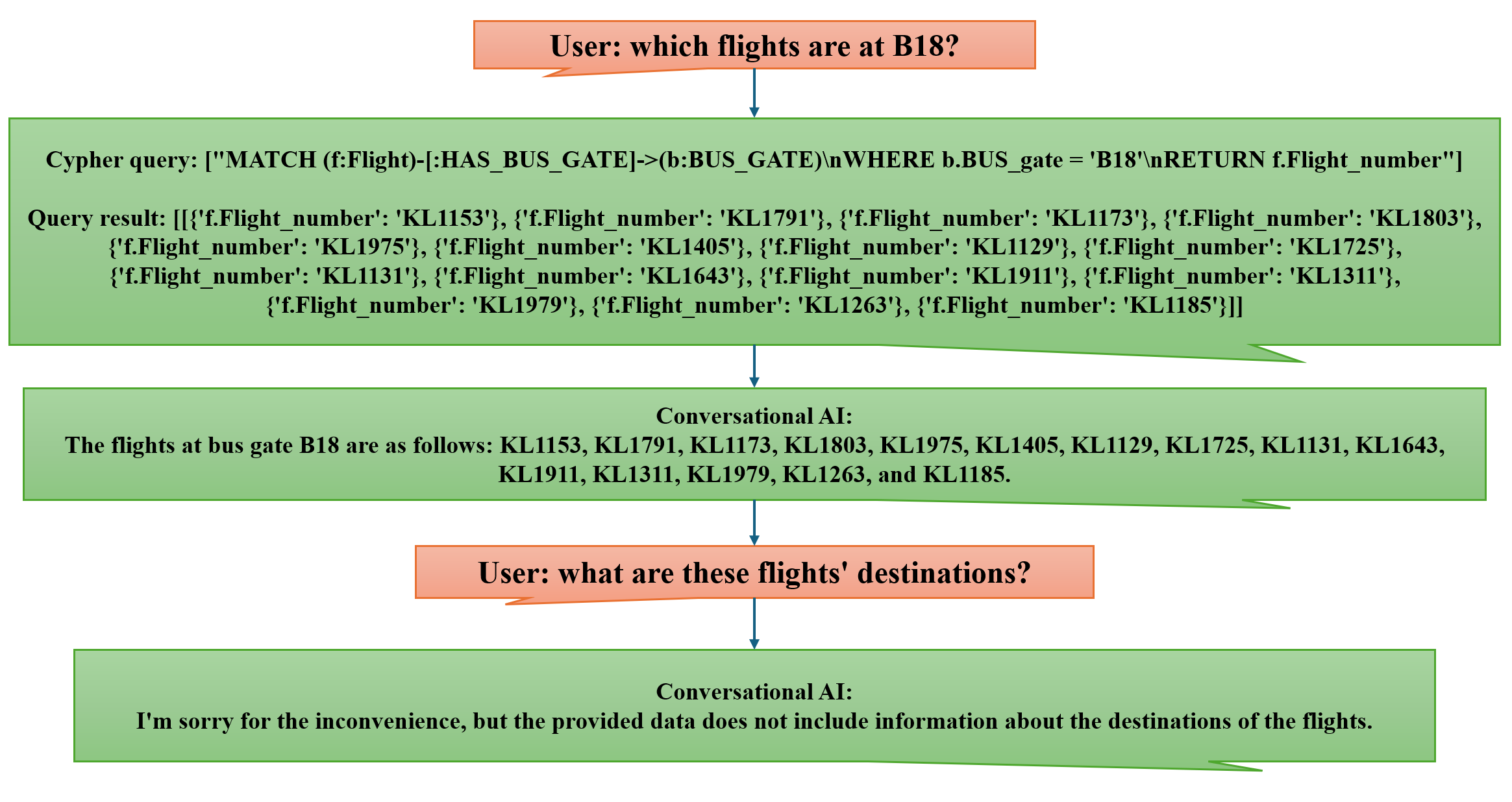}
    \caption{The same case for Graph RAG}
    \label{fig:halllucination_graph}
\end{figure*}

In contrast, for SQL RAG and Graph RAG, the retrieval process is more precise. In SQL RAG (Figure \ref{fig:halllucination_sql}), the natural language question is first converted into an SQL query that retrieves only the relevant information—in this case, flight numbers at gate B18. The results are then passed to the LLM to generate the final answer. Similarly, in Graph RAG (Figure \ref{fig:halllucination_graph}), a Cypher query retrieves only the flight numbers associated with gate B18. Since both SQL RAG and Graph RAG retrieve more targeted and accurate information compared to traditional RAG, they significantly reduce the likelihood of hallucinations.

It is important to note that hallucinations are not common even in traditional RAG, and they are not eliminated in SQL RAG or Graph RAG. Additionally, calculating the exact accuracy or rate of hallucinations across these RAG methods is challenging. However, because SQL RAG and Graph RAG retrieve information more accurately, they tend to reduce the occurrence of hallucinations compared to traditional RAG.

% \begin{figure*}[h!]
%     \centering
%     \begin{subfigure}[b]{0.5\textwidth}
%         \centering
%         \includegraphics[width=\textwidth]{latex/graph_rag_prompr.png}
%         \caption{Graph RAG Prompt}
%         \label{fig:graph_rag_prompt}
%     \end{subfigure}%
%     \begin{subfigure}[b]{0.5\textwidth}
%         \centering
%         \includegraphics[width=\textwidth]{latex/graph_rag_with_reasoning.png}
%         \caption{Graph RAG in Reasoning Questions}
%         \label{fig:graph_rag_reasoning}
%     \end{subfigure}
%     \caption{Graph RAG Analysis}
% \end{figure*}

% \subsubsection{Hallucination in Traditional RAG}
% Here shows more hallucination situation of Traditional RAG as the Figure \ref{fig:hallucination2}
  
% \begin{figure*}[h!]
%   \vspace{-5pt} %
%   \includegraphics[width=1\columnwidth]{latex/hallucination11.png}
%   \vspace{-10pt} 
%   \caption{Hallucination in the traditional RAG}
%   \label{fig:hallucination2}
% \end{figure*}

\end{document}